\def\year{2021}\relax
\documentclass[letterpaper]{article} 
\usepackage{aaai21}  

\usepackage{times}  
\usepackage{helvet} 
\usepackage{courier}  
\usepackage[hyphens]{url}  
\usepackage{graphicx} 
\usepackage{booktabs} 
\usepackage{multirow}
\usepackage{subfigure}
\usepackage{mathrsfs}
\usepackage[ruled,vlined]{algorithm2e}
\usepackage{algorithm2e}
\usepackage{graphicx}
\usepackage{balance} 
\usepackage{color}
\usepackage{amssymb}
\usepackage{amsmath}
\usepackage{enumitem}
\usepackage{natbib}  
\usepackage{caption} 
\usepackage{times}  
\usepackage{helvet} 
\usepackage{courier}  
\usepackage[hyphens]{url}  
\usepackage{graphicx} 
\usepackage{booktabs} 
\usepackage{multirow}
\usepackage{subfigure}
\usepackage{mathrsfs}
\usepackage[ruled,vlined]{algorithm2e}
\usepackage{algorithm2e}
\usepackage{graphicx}
\usepackage{balance} 
\usepackage{color}
\usepackage{amssymb}
\usepackage{amsmath}
\usepackage[switch]{lineno}  %
\usepackage{amsfonts}
\usepackage{subfig}
\usepackage{multirow}
\usepackage{eucal}

\title{AttnMove: History Enhanced Trajectory Recovery via Attentional Network}

\author {
    Tong Xia\textsuperscript{\rm 1},
    Yunhan Qi\textsuperscript{\rm 1},
    Jie Feng\textsuperscript{\rm 1},
    Fengli Xu\textsuperscript{\rm 1},
    Funing Sun\textsuperscript{\rm 2},
    Diansheng Guo\textsuperscript{\rm 2},
    Yong Li\textsuperscript{\rm 1}\thanks{This is the corresponding author.}
    \\
}
\affiliations {
    \textsuperscript{\rm 1}Beijing National Research Center for Information Science and Technology,\\
    Department of Electronic Engineering, Tsinghua University. \\
    \textsuperscript{\rm 2}Tencent Corporation. \\
    Emails: xia-t17@tsinghua.org.cn, liyong07@tsinghu.edu.cn
}

\begin{document}
\maketitle

\begin{abstract}
A considerable amount of mobility data has been accumulated due to the proliferation of location-based service. Nevertheless, compared with mobility data from transportation systems like the GPS module in taxis, this kind of data is commonly sparse in terms of individual trajectories in the sense that users do not access mobile services and contribute their data all the time. Consequently, the sparsity inevitably weakens the practical value of the data even it has a high user penetration rate. To solve this problem, we propose a novel attentional neural network-based model, named AttnMove, to densify individual trajectories by recovering unobserved locations at a fine-grained spatial-temporal resolution. To tackle the challenges posed by sparsity, we design various intra- and inter- trajectory attention mechanisms to better model the mobility regularity of users and fully exploit the periodical pattern from long-term history. We evaluate our model on two real-world datasets, and extensive results demonstrate the performance gain compared with the state-of-the-art methods. This also shows that, by providing high-quality mobility data, our model can benefit a variety of mobility-oriented down-stream applications.
\end{abstract}

\section{Introduction} \label{sec:introduction}
Widely adopted location-based services have accumulated large-scale human mobility data, which have great potential to benefit a wide range of applications, from personalized location recommendations to urban transportation planning~\cite{zheng2014urban}. Nevertheless, since users may not allow the service provider to collect their locations continuously, individual trajectory records in such data are extremely sparse and unevenly distributed in time, which inevitably harms the performance of downstream applications, even when it has a notable user penetration rate and covers a long period of time~\cite{gao2019privacy}. For example, with a limited number of trajectory records per day, it is difficult to predict an individual’s next location and recommend proper points-of-interests ~\cite{xi2019modelling}. As for collective mobility behaviour, because personal location records are missing for most of the time, it is also hard to produce exact estimates of hourly crowd flow in cities for emergency response~\cite{li2013efficient}. Therefore, it is of great importance to rebuild individual trajectories at a fine-grained spatial-temporal granularity by imputing missing or unobserved locations.

One commonest solution to this problem is to impute the missing value by treating individual trajectories as two-dimensional time series with latitude and longitude at each timestamp ~\cite{alwan1988time,moritz2017imputets}. As such, smoothing filters~\cite{alwan1988time,moritz2017imputets} and LSTM-based models~\cite{cao2018brits,wang2019deep} have been proposed. Their performance is acceptable when only a small percentage of locations are missing due to limited movement during a short time span. However, in highly sparse scenarios, their performance degrades significantly, since they fail to effectively model complex mobility regularity. Another line of study is to model users' transition regularity among different locations, so as to generate the missing locations according to the highest transition probability from observed location~\cite{liu2016predicting,wang2019deep}. But this kind of strategy is still insufficient in the sense that the observed records are unevenly distributed in time in LBS data, and thus transition regularity is incapable of inferring those locations which are continuously unobserved.

Luckily, human mobility has some intrinsic and natural characteristics like periodicity and repeatability, which can help to better rebuild the trajectory~\cite{gonzalez2008understanding,schneider2013unravelling,cho2011friendship}. In this regard, a promising direction is to leverage long-term mobility history, i.e., mobility records prior to the targeted trajectory to be recovered, considering both that daily movements are spatially and temporally periodic, and that collecting long-term data is quite achievable in LBS. In addition, previous work has also shown that explicitly utilizing historical trajectory can help with next location prediction~\cite{feng2018deepmove}, which is similar to our task. All of this has inspired us to design a history enhanced recovery model.

However, trajectory recovery is still challenging for the following reasons: first of all, the high sparsity of both targeted day and historical trajectories hinder us from inferring the missing locations by spatial-temporal constraints, i.e., how far a user can move in the unobserved periods. Because of the high uncertainty between two consecutive records in a sparse trajectory, a framework which can better model the mobility pattern and reduce the number of potentially visited locations is needed. The second is how to distill periodical features from huge historical data effectively, considering that real-world historical data includes a great deal of noise. Researchers have proposed detecting the locations of the home and the workplace from historical trajectories, in order to build the basic periodic pattern. This is straightforward but insufficient because other locations are neglected~\cite{chen2019complete}. Another way is to directly exploit the most frequently visited location at the targeted time slot from multiple historical trajectories as imputation~\cite{liu2016predicting}. However, historically more popular locations may not be the ones which are missing on any targeted day because mobility is influenced by many factors and thus some locations can only be visited occasionally. Thus, deciding how well to rely on history is the third challenge.

Keeping the above challenges in mind, we aim to propose a novel attentional neural network-based mobility recovery model named \textbf{AttnMove}, which is dedicated to handling sparsity and exploiting long-term history.  For clarity, we define the targeted day's trajectory as the current trajectory and any trajectories before the targeted day as historical trajectories. The proposed AttnMove model can be broken down into three key components, which address the main challenges accordingly.  Firstly, to capture the mobility pattern and indicate the most likely visited areas for the missing records, we design a \emph{current processor} with \emph{intra-trajectory attention} mechanism to initially fill in the blanks of the current trajectory. We choose attention mechanism rather than RNN structure for the reason that sparse trajectories have few sequential characteristics and all the observed locations should be considered equally regardless of the visited order. For example, a user's locations at 6 AM, 9 AM, 3 PM and 6 PM are observed: to recover the location at 9 PM, we should give priority to the location at 6 AM and the adjacent area (i.e., home area) instead of at 6 PM (i.e., commuting time). For the second challenge, we design a \emph{history processor} with another \emph{intra-trajectory attention} to distill periodical features from multiple historical trajectories. By aggregating, more information from the long-term history can be leveraged for recovery. Finally, to fuse the features from current and history processors and generate missing locations, we propose a \emph{trajectory recovery} module with \emph{inter-trajectory attention} and \emph{location generation attention} mechanisms. To be specific, the former mechanism yields attention weights to select locations from history based on current mobility states, and the later further considers spatial-temporal constraints from observed locations to better rebuild the trajectory.

Overall, our contributions can be summarized as follows:
\begin{itemize}
    \item We study the problem of rebuilding individual trajectories with a fine-grained spatial-temporal granularity from sparse mobility data by explicitly exploiting long-term history, which brings high-quality mobility data to facilitate a wide range of down-stream applications.
    \item We propose a novel attentional neural network model AttnMove for missing location recovery, behind which the key idea is to utilize attention mechanism to model mobility regularities and extract periodical patterns from multiply trajectories of different days, and then fuse them to rebuild the trajectories.
    \item We conduct extensive experiments on two real-life mobility datasets. Results show that our AttnMove model significantly outperforms state-of-the-art baselines in terms of improving recovery accuracy by 4.0\%$\sim$7.6\%\footnote{Codes are  available in https://github.com/XTxiatong/AttnMove}. 
\end{itemize}

\section{Preliminaries} \label{sec:problem}
In this section, we  introduce the definition and notations we use in this paper.
\vspace{2mm}
\\\textbf{Definition 1} (\textbf{Trajectory}). \emph{We define a trajectory as a user's time-ordered location sequence in one day.
Let $\mathcal{T}^n_u: l^{n,1}_u \rightarrow l^{n,2}_u... \rightarrow l^{n,t}_u...\rightarrow l^{n,T}_u$ denote user $u$'s $n$-th day's trajectory,  where $l^{n,t}_u$ is the location of $t$-th time slot for a given time interval (e.g., every 10 minutes). Note that if the location of time slot $t$ is unobserved, $l^{n,t}_u$ is denoted by null, named missing location. }
\vspace{2mm}
\\\textbf{Definition 2} (\textbf{Current and Historical Trajectory}). \emph{Give a targeted day $n$ and user $
u$'s trajectory $\mathcal{T}^n_u$, we define $\mathcal{T}^n_u$ as the user's current trajectory, and the historical trajectories are defined as $u$'s trajectories in the past days, i.e., $\{\mathcal{T}^1_u,\mathcal{T}^2_u,...,\mathcal{T}^{n-1}_u\}$}.

\vspace{2mm}
When most of the locations in current day's trajectory $\mathcal{T}^n_u$ are missing, exploiting history is beneficial for the recovery. For this reason, we are motivated to formulate the investigated problem as:
\vspace{2mm}
\\\textbf{Problem} (\textbf{History Enhanced Trajectory Recovery}). \emph{Given user $u$'s trajectory $\mathcal{T}_u^n$ with the historical trajectories  $\{\mathcal{T}^1_u,\mathcal{T}^2_u,...,\mathcal{T}^{n-1}_u\}$, recover the missing locations, i.e., $\forall$ null in $\mathcal{T}_u^n$, to rebuild the current day's complete trajectory. }

\begin{figure}[h]
    \centering
    \includegraphics[width=0.48\textwidth]{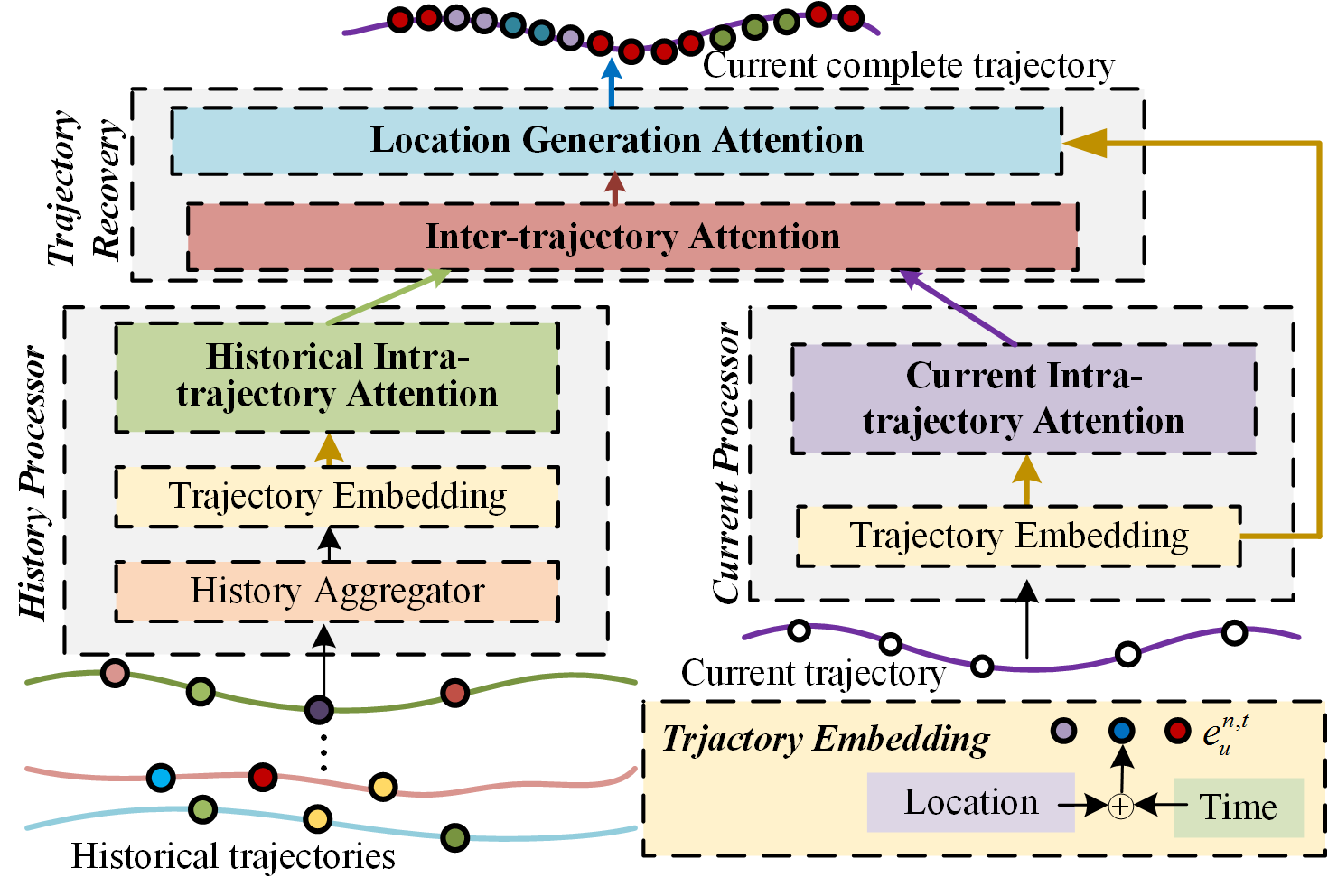}
    \caption{Main architecture of the proposed AttnMove model, where current and historical trajectories are first processed separately, and then are fused to generate locations for missing time slots.}
    \label{fig:main}
\end{figure}

\section{AttnMove: Model Design} \label{sec:method}
To solve the above defined problem, we devise a novel model AttnMove, which first processes historical and current day's trajectories separately, and then integrates their features to comprehensively determine the location to be recovered. The architecture of AttnMove is illustrated in Fig.~\ref{fig:main}.
In our AttnMove, to projects sparse location and time representations into dense vectors which are more expressive and computable, a \emph{trajectory embedding module} is designed as a pre-requisite component for other modules.
Then, in order to extract periodical patterns, multiply historical trajectories are feed into a \emph{history processor} to be aggregated.  Paralleled with \emph{history processor} is a \emph{current processor}. We design it to enhance spatial-temporal dependence to better model mobility regularities.
Finally, to fuse historical and current trajectories from the above modules and  generate locations as recovery, a \emph{trajectory recovery module} is proposed as the finally component. In the following, we elaborate on the details of those four modules.

\subsection{Trajectory Embedding Module}
To represent the spatial-temporal dependency, we jointly embed the time and location into dense representations as the input of other modules. Specifically, for each location $l \in \mathcal{L}$ including the missing one \emph{null}, we set up a trainable embedding vector $e_l \in  \mathbb{R}^{d}$. All location embeddings are denoted as a matrix $E_l \in \mathbb{R}^{|\mathcal{L}+1| \times d}$. As for mobility modeling, temporal information is also important, therefore we also set up embeddings for time slots. Following \cite{vaswani2017attention}, for each time slot $t$, we generate its embedding as follows,
\begin{equation}
\begin{cases}
e_t(2i) = \sin(t/10000^{2i/d}), \ \ \ \\
e_t(2i+1) = \cos(t/10000^{2i/d}),
\end{cases}
\end{equation}where $i$ denotes the $i$-th dimension. The time embedding vectors have the same dimension $d$ with location embedding.
Finally, for each time slot $t$, we sum the time and location embedding vectors into a single one, denoted by $e_u^{n,t} \in \mathbb{R}^{d}$ as follows,
\begin{equation}
\label{equ:embed}
   e_u^{n,t}=e_l+e_t,
\end{equation}which also benefits the follow-up computation as a lower dimensional vector than the original one-hot ones.

\subsection{History Processor} 
When exploiting the historical trajectory, information provided by a single historical trajectory with one day's observed locations is insufficient to capture periodical characters as each day's records are sparse. To utilize multiple trajectories jointly, we design a \emph{\textbf{history aggregator}} as follows,
\begin{equation}
\mathcal{T}^p_u = \mathcal{T}^1_u \oplus \mathcal{T}^2_u\cdot \cdot \cdot \oplus\mathcal{T}^{n-1}_u,
\end{equation}where $\oplus$ means to extract the location with highest visiting frequency in the corresponding time slot. Then, $\mathcal{T}^p_u$ is embedded to $e^p_u$ with each time slot represented by $e^{p,t}_u$ according to \eqref{equ:embed}. 
As such, $\mathcal{T}^p_u$ is less sparse than any single historical trajectory.
However, as it is possible that user $u$ did not generate any record at a certain time slot previously, locations can still be missing after aggregating. Therefore, we further propose the following  mechanism to infill  $\mathcal{T}^p_u$:
\vspace{2mm}
\begin{figure}[t]
    \centering
    \includegraphics[width=0.45\textwidth]{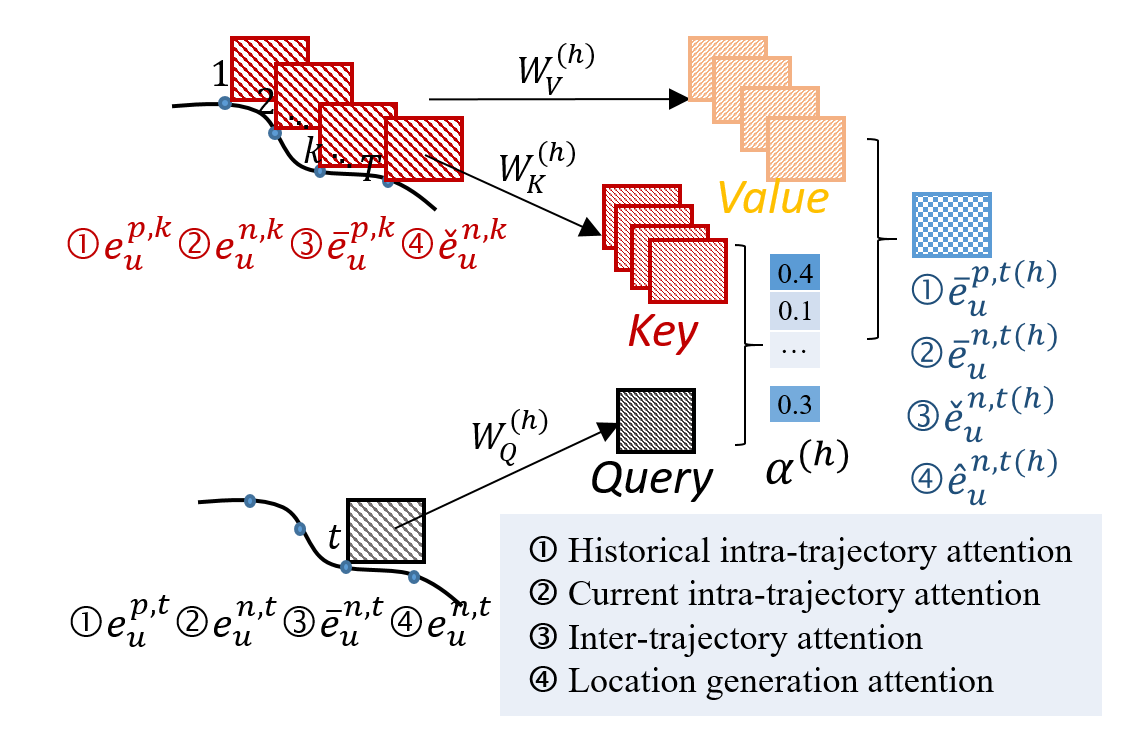}
    \caption{The architecture of proposed trajectory attention mechanisms under one head, where the new representation is a combination of embeddings of value time slots conditioned on attention weights $\alpha^{(h)}_{t,k}$ from query and key time slots. Note that superscript $p$ denotes history while $n$ is for the current trajectory. }
    \label{fig:4atten}
    \vspace{-5mm}
\end{figure}
\\\emph{\textbf{Historical Intra-trajectory Attention}}.
Intuitively, the missing location can be identified by its time interval and distance to other observed locations. Take a simple example,  $l^{p,4}_u$ is most likely in the geo-center of $l^{p,3}_u$ and $l^{p,5}_u$. In the embedding space, this can be expressed by $e^{p,4}_u = 0.5e^{p,3}_u + 0.5e^{p,5}_u $. However, the actual scenario would be much more complex than this as people do not move uniformly. Therefore, we use a multi-head attentional network to model the spatial-temporal relation among trajectory points. Specifically, we define the correlation between time slot $t$ and $k$ under head $h$ as follows,
\begin{align}
\label{equ:atten1}
&\alpha^{(h)}_{t,k} = \frac{\exp(\phi^{(h)}(e_u^{p,t},e_u^{p,k}))}{\sum_{g=1}^{T}\exp(\phi^{(h)}(e_u^{p,t},e_u^{p,k}))},\\
\label{equ:atten2}
&\phi^{(h)}(e_u^{p,t},e_u^{p,k}) = \langle W_Q^{1(h)}e_u^{p,t}, W_K^{1(h)}e_u^{p,k} \rangle,
\end{align}where  $W_Q^{1(h)},W_K^{1(h)} \in \mathbb{R}^{d'\times d}$ are transformation matrices and $\langle,\rangle$ is the inner product function. Next, we generate the representation of time slot $t$ under each head via combining all locations in other time slots guided by coefficient  $\alpha^{(h)}_{t,k}$:
\label{equ:atten3}
\begin{equation}
    \widetilde{e}_u^{p,t(h)} = \sum_{k=1}^T \alpha^{(h)}_{t,k}(W_V^{1(h)}e_u^{p,k}),
\end{equation}where $W_V^{1(h)} \in \mathbb{R}^{d'\times d}$ is also a transformation matrix.
 Furthermore, we make use of different combinations which models different spatial-temporal dependence and collect them as follows,
\label{equ:atten4}
\begin{equation}
    \widetilde{e}_u^{p,t} = \widetilde{e}_u^{p,t(1)}\parallel\widetilde{e}_u^{p,t(2)}\parallel \cdot\cdot\cdot \parallel\widetilde{e}_u^{p,t(H)},
\end{equation}where $\parallel$ is the concatenation operator, and $H$ is the number of total heads. 
To preserve the representation of raw locations, we add a standard residual connections in the network. Formally,
\label{equ:atten5}
\begin{equation}
    \overline{e}_u^{p,t} =ReLU(\widetilde{e}_u^{p,t} + W^1{e}_u^{p,t}),
\end{equation}where $W^1 \in \mathbb{R}^{d'H\times d}$ is the projection matrix in case of dimension mismatching, and $ReLU(z)=\max(0,z)$ is non-linear activation function.

With the historical intra-trajectory attention layer, the representation of history trajectory $e_u^p$ is updated into a more expressive form $\overline{e}_u^{p}$, which is shown in Fig. \ref{fig:4atten}. We can stack multiple such layers with the output of the previous layer as the input of the next one. By doing this, we can extract more complex mobility features from historical trajectories.

\subsection{Current Processor} 
When recovering the trajectory for user $u$, it is necessary to consider the locations visited before and after the missing time slots, which enclosure the geographical area of the missing locations. Since locations can be missed for several consecutive time slots, the spatial constraint is weak. Therefore, we conduct intra-trajectory attention on the current trajectory $\mathcal{T}_u^n$ to capture the current day's mobility pattern:
\vspace{1mm}
\\\emph{\textbf{Current Intra-trajectory Attention}}.
 The first step is to embed $\mathcal{T}_u^n$ into a dense representation $e_u^n$ via the aforementioned mobility embedding module. Next, we conduct an attention mechanism on $e_u^n$, which has same network structure as given in (4)-(8) 
with input history pattern $e_u^p$ replaced by current trajectory representation $e_u^n$. We denote the relevant projection matrices as $W_Q^2, W_V^2,W_K^2,W^2$, respectively.  We also stack this layer for multiple times to fully capture the spatial-temporal correlation. Then, after updating, we obtain an enhanced current trajectory represented by $\overline{e}_u^{n,t}$.

\subsection{Trajectory Recovery Module}
After extracting the historical and current features, the problem is how much to depend on the interpolation via current observed locations or rely on the candidates generated from historical trajectories. Intuitively, a good solution is to compare the current mobility status with historical one, and combine the history information with current interpolation results according to their similarity. To achieve this, we propose the follow mechanism:
\vspace{2mm}
\\\emph{\textbf{Inter-trajectory Attention}}.
We define the similarity of current and historical trajectory denoted by $\alpha$ as the correlation between the enhanced representation in corresponding time slots, i.e.,  between $\overline{e}_u^{n,t}$ and $\overline{e}_u^{p,k}(\forall t,k=1,2,...,|T|)$. Then we combine history candidates by the similarity $\alpha$, followed by a residential connection to remain the raw interpolation results. 
Finally, the fused trajectory $\check{e}_u^{n}$ is generated from $\overline{e}_u^{n,t}$ and $\overline{e}_u^{p,k}$, which can be expressed as follows,
\begin{align}
\label{equ:atten6}
&\alpha^{(h)}_{t,k} = \frac{\exp(\phi^{(h)}(\overline{e}_u^{n,t},\overline{e}_u^{p,k}))}{\sum_{g=1}^{T}\exp(\phi^{(h)}(\overline{e}_u^{n,t},\overline{e}_u^{p,g}))},\\
&\phi^{(h)}(\overline{e}_u^{n,t},\overline{e}_u^{p,k}) = \langle W_Q^{3(h)}\overline{e}_u^{n,t}, W_K^{3(h)}\overline{e}_u^{p,k} \rangle,\\
&\widetilde{e}_u^{p,t(h)} = \sum_{k=1}^T \alpha^{(h)}_{t,k}(W_v^{3(h)}\overline{e}_u^{p,k}),\\
&\widetilde{e}_u^{p,t} = \widetilde{e}_u^{p,t(1)}\parallel\widetilde{e}_u^{p,t(2)}\parallel \cdot\cdot\cdot \parallel\widetilde{e}_u^{p,t(H)},\\
&\check{e}_u^{n,t} =ReLU(\widetilde{e}_u^{p,t} + W^3\overline{e}_u^{n,t}),
\end{align}where $W_Q^{3(h)}, W_K^{3(h)}, W_V^{3(h)}, W^3$ are the projection matrices.

With the fused trajectory that contains both history mobility information and current spatial-temporal dependence, we are ready to recovery the missing location. We use the following mechanism to generate the representation of the missing location, and then use it to identify the specific location.
\vspace{2mm}
\\\emph{\textbf{Location Generation Attention}}.
To generate the final representation denoted by $\hat{e}_u^{n,t}$, we define a temporal similarity among the current trajectory represented by $e_u^n$ and the fused trajectory represented by $\check{e}_u^{n,t}$ as $\alpha$, which can be derived by (9)-(10) with $\overline{e}_u^{n,t}$ and $\overline{e}_u^{p,k}$ replaced by $\check{e}_u^{n,t}$ and $e_u^{n,t}$, respectively. Then, $\hat{e}_u^{n,t}$ for time slot $t$ is a combination of $\check{e}_u^{n,t}$ according to $\alpha$, which is the same as (11)-(13). The projection matrices are denoted by $W_Q^4, W_V^4,W_K^4,W^4$, respectively.

Once we obtain $\hat{e}_u^{n,t}$, we are able to compute the probability that user $u$  visits location $l$ at time slot $t$ as follows,
\begin{equation}
    P_u^{n,t}(l) =\frac{\langle \hat{e}_u^{n,t}, e_l \rangle} {\sum_{k\in \mathcal{L}} \langle\hat{e}_u^{n,t}, e_k\rangle},
\end{equation}where $\langle,\rangle$ is the inner product function, and $P_u^{n,t} \in \mathbb{R}^{|\mathcal{L}|}$ denotes the normalized probabilities of all location visited  at time slot $t$.
In practice, the location with maximum probability is identified as the missing location.

\subsection{Training}
Overall, the parameters of AttnMove include projection matrices and location embedding matrix, denoted by $\theta = \{ E_l, W^i_Q,W^i_V,W^i_K,W^i, i=1,2,3,4 \}$. 
To train the model,  we use \emph{cross entropy} as the loss function:
\begin{equation}
 L(\theta) = -\sum_{u\in \mathcal{U}}\sum_{n \in \mathcal{N}} \sum_{t \in \mathcal{T^M}} \langle y_u^{n,t}, \log( P_u^{n,t})\rangle + \lambda \|\theta\|^2,
\end{equation}where $\langle,\rangle$ is the inner product, $y_u^{n,t}$ is the one-hot representation of user $u$'s  location in $n$-th day's $t$-th time slot, $\mathcal{T^M}$ denotes the missing time slots, and $\lambda $ is a parameter to control the power of regularization.  Training algorithm is illustrated in Algorithm~\ref{train}, and the process is done through stochastic gradient descent over shuffled mini-batches across Adam optimizer~\cite{kingma2014adam}.
In addition, our model is implemented by Python and Tensorflow~\cite{abadi2016tensorflow}. We train our model on a linux server with a TITAN Xp GPU (12 G Memory) and a Intel(R) Xeon(R) CPU @ 2.20GHz. 

\vspace{-2mm}
\begin{algorithm}[h]
    \DontPrintSemicolon
    \caption{Training Algorithm for AttnMove}
    \label{train}
    \textbf{Input:} Trajectory sets $\{\mathcal{T}_1,\mathcal{T}_2,...,\mathcal{T}_U\}$; \\
    \textbf{Output:} Trained Model $\theta$.\\
    //construct training instances: $\mathcal{D} \longleftarrow \emptyset$ \\
    \For{$u$ $\in\{u_1,...,\mathcal{U}\}$}
    {
        \For{$n$ $\in\{1,2,...,\mathcal{N}\}$}{
            Put a training instance $(u,n,\mathcal{T^M})$ into $\mathcal{D}$
        }
    }
    // Train the model: initialize the parameters $\theta$ \\
    \For{$i$ $\in\{1,2,...,EPOCH\}$}{
        Select one batch $\mathcal{D}_b$ from $\mathcal{D}$;\\
        Update $\theta$ by minimizing the objective $ L(\theta)$ with $\mathcal{D}_b$;\\
        Stop training when criteria is met;\\
    }
    Output trained model $\theta$
\end{algorithm}

\section{Performance Evaluation} \label{sec:evaluation}
\subsection{Datasets} 
\begin{itemize}[leftmargin=10pt]
\setlength{\itemsep}{0pt}
\setlength{\parsep}{0pt}
\setlength{\parskip}{0pt}
 \item \textbf{Tencent}\footnote{https://lbs.qq.com/}: This data is collected from the most popular social network and location-based service vendor Tencent in Beijing, China from June 1st$\sim$30th, 2018. It records the GPS location of users whenever they request the localization service in the applications.
 \item  \textbf{Geolife}\footnote{https://www.microsoft.com/en-us/research/project/geolife-building-social-networks-using-human-location-history/}: This open data is collected from Microsoft Research Asia Geolife project by 182 users from April 2007 to August 2012 over all the world. Each trajectory is represented by a sequence of time-stamped points, containing longitude and altitude \cite{zheng2010geolife}.
\end{itemize}

\begin{table}[h]
\centering
\vspace{-2mm}
\caption{Basic statistics of mobility datasets.}
\vspace{-2mm}
    \resizebox{0.48\textwidth}{!}{%
    \begin{tabular}{@{}c|cccccc@{}}
    \toprule
    \textbf{Dataset} & \textbf{City} & \textbf{Duration} &  \textbf{\#Users} & \textbf{\#Traj.} & \textbf{\#Distinctive Loc.}\\ \midrule
     Tencent   &  Beijing      &  1 month & 4265  & 39,422 & 8,998  \\
     Geolife   &  Beijing      &  5 years      & 40    & 896   & 3,439 \\ \bottomrule
    \end{tabular}}
\label{tab:dataset}
\end{table}

\textbf{Pre-processing:} To represent the location, we crawl the road network of Beijing from online map$^2$, and divide the area into 10,655 blocks. Each blocks is treated as a distinctive location with the size of 0.265 $km^2$ on average. 
Following~\cite{chen2019complete}, we set time interval as 30 minutes for both two datasets. For model training and testing, we filter out the trajectories with less than 34 time slots (i.e., 70\% of one day) and the users with less than 5 day's trajectories for Tencent, and filter out the trajectories with less than 12 time slots and the users with less than 5 day's trajectories for Geolife.  The final detailed statics are summarized in Table \ref{tab:dataset}.

\subsection{Baselines}
 We compare our AttnMove with  several representative baselines. Among them, the first four are directly based on our knowledge about human mobility regularity and the last four are the state-of-the-art deep learning models which can extract more complex mobility features:
\begin{itemize}[leftmargin=10pt]
\setlength{\itemsep}{0pt}
\setlength{\parsep}{0pt}
\setlength{\parskip}{0pt}
    \item \textbf{Top} \cite{liu2016predicting}: It is a simple counting-based method. The most popular locations in the training set are used as recovery for each user.
    \item \textbf{History} \cite{li2019reconstruction}: In this method, the most frequently visited locations of each time slot in historical trajectories are used for recovery.
    \item \textbf{Linear}  \cite{hoteit2014estimating}: This is also a rule-based method. It recovers the locations by assuming that users are moving straightly and uniformly.
    \item \textbf{RF} \cite{li2019reconstruction}: RF is a feature-based machine learning method. Entropy and radius of each trajectory, the missing time slot, the location before and after the missing time slot are extracted as features to train a random forest classifier for recovery.
    \item \textbf{LSTM}  \cite{liu2016predicting}: It models the forward sequential transitions of mobility by recurrent neural network, and use the prediction for next time slot for recovery.
    \item \textbf{BiLSTM}  \cite{zhao2018prediction}: It extends LSTM by bi-directional RNN to consider the spatial-temporal constraints given by all observed locations.
    \item \textbf{DeepMove} \cite{feng2018deepmove}: Besides modeling sequential transitions, DeepMove incorporates the historical trajectories by attention mechanism for next location prediction. We also use the prediction result for recovery.
    \item \textbf{Bi-STDDP}  \cite{xi2019modelling}: This is the latest missing location recovery method, which jointly models user preference and the spatial-temporal dependence given the two locations visited before and after the targeted time slot to identify the missing ones.
\end{itemize}

Apart from those baselines, to evaluate the effectiveness of our designed mechanism to exploit history, we also compare \textbf{AttnMove} with its simplified version \textbf{AttnMove-H}, where history processor and inter-trajectory attention layer are removed.

\subsection{Experimental Settings}
To evaluate the performance, we mask some time slots as ground-truth to recover. Since about 20\% locations are missing in the raw datasets on average,  we randomly mask 30 and 10 time slots per day for Tencent and Geolife. 
We sort each user's trajectories by time, and take the first 70\% as the training set from the fourth day (to guarantee that each trajectory has at least three days as history), the following 10\% as the validation set and the remaining 20\% as the test set. Linear, Top and History are individual models, while other models are shared by the users in one dataset. The regularization factor $\lambda $ is set as 0.01. 

We employ the widely used metrics \emph{Recall} and Mean Average Precision (\emph{MAP})~\cite{wang2019deep,liu2016predicting}. \emph{Recall} is 1 if the ground-truth location is recovered with maximum probability; otherwise is 0. The final \emph{Recall} is the average value over all instances. \emph{MAP} is a global evaluation for ranking tasks, so we use it to evaluate the quality of the whole ranked lists including all locations.  The larger the value of those two metrics is, the better the performance will be. We aslo make use of the metric of \emph{Distance}, which is the geographical distance between the center of recovered location and the ground-truth. The smaller the \emph{Distance} is, the better the performance will be.

 \subsection{Experiment Results}
\begin{table}[t]
    \centering
    \caption{Overall performance comparison. The best result in each column is in bold, while the second is underlined.}
    \resizebox{0.5\textwidth}{!}
  {%
    \begin{tabular}{ccccccc}
        \toprule  
        \multirow{2}{*}{\textbf{Model}}  & \multicolumn{3}{c}{\textbf{Tencent}}    & \multicolumn{3}{c}{\textbf{Geolife}}  \\ \cmidrule(r){2-4} \cmidrule(r){5-7} 
                      & \emph{Recall}  & \emph{MAP}  & \emph{Dis.(m)} &\emph{Recall}& \emph{MAP}   & \emph{Dis.(m)}   \\ \midrule
        TOP            & 0.5879  & 0.6123  & 2530      & 0.2757   & 0.2879         & 5334     \\
        History        & 0.4724  & 0.4937  & 1613      & 0.2505   & 0.2648         & 5116   \\
        Linear         & 0.6234  & 0.6567  & 1145      & 0.3642   & 0.3889         & \textbf{2383}   \\ 
        RF             & 0.4848  & 0.4912  & 8128      &0.2551   & 0.2540         & 6144     \\ \midrule
        LSTM           & 0.6084  & 0.6722  & 3759      &0.2725   & 0.3314    & 5864      \\
        BiLSTM         & 0.7090  & 0.7805  & 1371      &0.3471   & 0.4148    & 5097      \\
        DeepMove       & \underline{0.7259}  & \underline{0.7872}  & 1322      &0.3391   & 0.4015    & 4912     \\
        Bi-STDDP       & 0.7037  & 0.7831  & \underline{1168}      &\underline{0.3701}   & \underline{0.4510}    & 4041  \\\midrule
        AttnMove-H     & 0.7358  & 0.7999  & 1083      &0.3853   & 0.4501    & 4129      \\ 
        AttnMove  & \textbf{0.7646} & \textbf{0.8249} & \textbf{934}  & \textbf{0.3982} & \textbf{0.4691}    & \underline{3886}   \\\bottomrule 
    \end{tabular}}
    \label{tab:result}
\end{table}

\subsubsection{Overall Performance}
We report the overall performance in Table~\ref{tab:result} and have the following observations:

1) Rule-based models fail to achieve high accuracy with \emph{Recall} and \emph{MAP} lower than 0.66 in Tencent and lower than 0.39 in Geolife. Although intuitively, frequently visited locations, historically visited locations and moving constraints are helpful for recovery, simply utilizing them cannot achieve favourable performance because these methods cannot model complex mobility regularity. 

2) RNN-based models are  insufficient but bidirectional ones can  reduce the \emph{Distance} error. A plausible reason is spatial-temporal constraints from all the observed locations is crucial instead of the sequential regularity for recovery. In addition, our AttnMove achieves further performance gain over the best RNN methods. It indicates the superiority of attention mechanisms. 

3) AttnMove outperforms all the baselines for all the metrics except for \emph{Distance} in Geolife. Specifically, \emph{Recall} and \emph{MAP} of AttnMove outperforms the best baselines by 4.0\%$\sim$7.6\%. \emph{Distance} is also reduced by 20\%  in Tencent. It is possible that Geolife is a small dataset for training, therefore \emph{Distance} of AttnMove cannot outperform Linear. But higher \emph{MAP} of AttnMove shows that although the generated location with the highest probability does not hit the ground-truth, which leads to higher \emph{Distance}, the correct location is on the top of recovered locations list.

 4) When comparing DeepMove with LSTM, and AttnMove with AttnMove-H, we can observe that history plays its role in reducing the uncertainty and improving recovery accuracy.  

To conclude, our AttnMove model achieves preferable results compared with both rule-based and machine learning-based methods. This justifies our model's effectiveness in capturing mobility pattern and exploiting history, and therefore it is a powerful model to rebuild fine-grained trajectories.

\begin{figure}[h]
  \centering
    \vspace{-3mm}
  \subfigure[Location embedding with color denoting cluster.]{
  \includegraphics[width=0.21\textwidth]{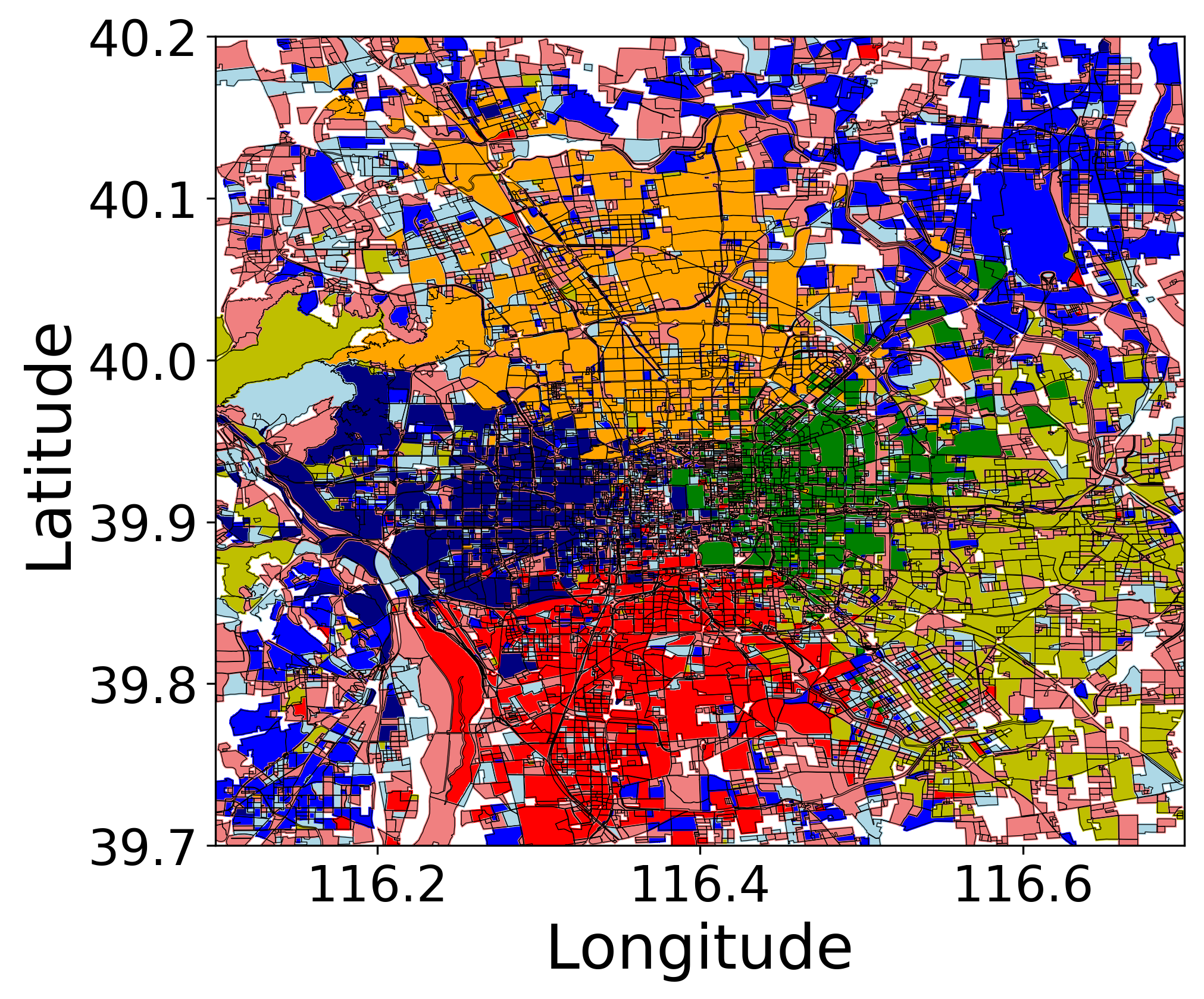}}
  \subfigure[Attention weight.]{
  \includegraphics[width=0.22\textwidth]{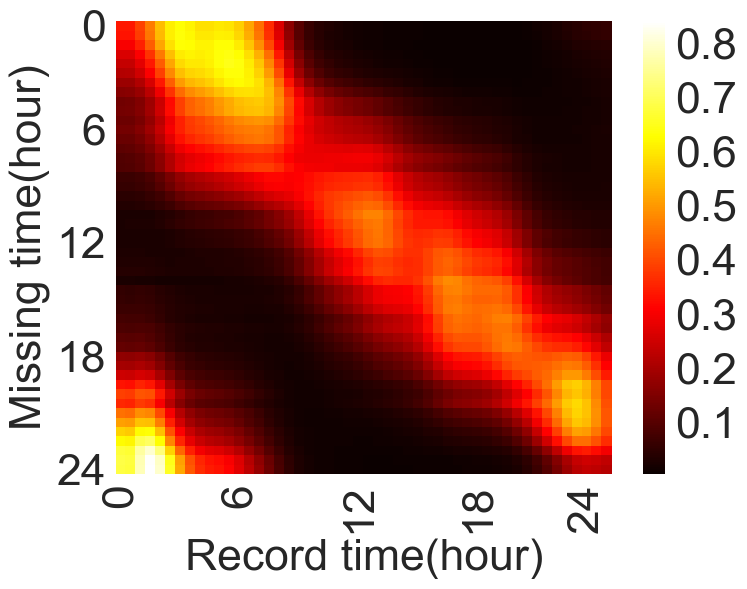}}
  \vspace{-4mm}
  \caption{Visualization results.}
    \vspace{-4mm}
\end{figure}

\subsubsection{Visualization Analysis}
Besides the above quantitative analysis, we also make some visualization study.
First, to investigate whether our learned trajectory embedding can learn the spatial correlation automatically,  we cluster the regions by using their embedding $e_l$ as features via k-means with Euclidean distance. Figure3 (a) shows their geographic distribution with clustering results. We can observe that adjacent locations generally share the same color indicating they are also closed in the embedding space, demonstrating that the spatial adjacent relation has been modeled.
Second, we visualize the attention weights of each head in each layer for different users, and present the average attention weights of all trajectories in the final current intra-trajectory attention layer in Figure 3(b). Overall, the diagonal entries are highlighted indicating that the target location more likely depends on the locations of adjacent time slots. More importantly, the bright part in bottom-left corner indicates apart from adjacent time, morning and night locations are highly related as well.  It demonstrate that our attentional network can movement regularity and periodical characters simultaneously.

\subsubsection{Ablation Analysis}
We analyze the effects of each trajectory attention mechanism. We create ablations by removing them one by one, i.e., using the embedding of the corresponding time slot directly instead of using a weighted combination. We report the results in Table~\ref{tab:result_a}. As expected, AttnMove outperforms all the ablations, indicating each attention can improve recovery performance. Specifically, when removing current intra-trajectory attention, the performance declines most significantly. This is because the attention can effectively strengthen the spatial constraints for the missing locations, without which even utilizing history, the improvement is limited.
\begin{table}[h]
   \vspace{-2mm}
    \centering
    \caption{Impact of attention mechanisms on Tencent dataset, where $\delta$ denoted the performance decline.}
    \vspace{-2mm}
    \resizebox{0.5\textwidth}{!}
  {
    \begin{tabular}{c|ccc}
        \toprule  
      Ablation                              & \emph{Recall($\Delta$)}  & \emph{MAP($\Delta$)}     & \emph{Dis.($\Delta$)(m)}   \\ \midrule
      Historical intra-trajectory attention   & 0.7457(2.5\%)  & 0.8074(2.1\%)  & 986(6.3\%)  \\\midrule
      Current intra-trajectory attention      & 0.7002(8.4\%)  & 0.7793(5.6\%)  & 1636(75.2\%)  \\\midrule
      Inter-trajectory attention   & 0.7434(2.8\%)  & 0.8119(1.6\%)  & 1063(13.8\%)  \\\midrule
      Location generation attention      & 0.7552(1.2\%)  & 0.8152(1.2\%)  & 962(3.0\%)  \\\bottomrule
    \end{tabular}}
    \label{tab:result_a}
\end{table}

\begin{table}[t]
\vspace{-2mm}
\caption{Performance \emph{w.r.t} missing ratios on Tencent dataset.}
\vspace{-2mm}
\resizebox{0.48\textwidth}{!}{%
\begin{tabular}{c|c|c|c|c|c}
\toprule
\multicolumn{2}{c|}{\textbf{Missing Rate}} & \textbf{60\%-70\%} & \textbf{70\%-80\%} & \textbf{80\%-90\%} & \textbf{90\%-100\%} \\ \hline
\multicolumn{2}{c|}{Precentage}             & 19.2\%             & 30.2\%             & 41.8\%             & 8.8\%               \\ \hline
\multirow{2}{*}{BiSTDDP}        & \emph{Recall}    & 0.7637             & 0.7366             & 0.6636             & 0.6501              \\ \cline{2-6} 
                                 & Distance(m)  & 755                & 1022               & 1358               & 1667                \\ \hline
\multirow{2}{*}{AttnMove}        & \emph{Recall}    & 0.8064             & 0.7820             & 0.7465             & 0.6993              \\ \cline{2-6} 
                                 & Distance(m)  & 651                & 848                & 1050               & 1294                \\ \bottomrule
\end{tabular}%
}\label{tab:rate}

\end{table}

\subsubsection{Robustness Analysis}
We also conduct experiments to evaluate the robustness of AttnMove when applied in different scenarios.
Firstly, we study the recovery performance w.r.t missing ratio,i.e., the percentage of missing locations. The results is presented in Table \ref{tab:rate}.
 As missing ratio rises from 50\% to almost 100\%, it is more intractable to recover correctly, while our model can maintain a significant gap (i.e., more than 3\%) with the state-of-the-art baseline. 
Second, we present the performance in different of a day and of trajectories with a different number of locations visited. As we can observe from Figure 4, during the day time when people are more likely to move or for these trajectories where numerous locations are visited, it is more difficult to recovery correctly and thus \emph{Recall} declines. Nevertheless, our \emph{Recall} can always outperform baselines by more than 5\%. Those results demonstrate the robustness of our proposed model in different scenarios.

\begin{figure}[t]
\vspace{-1mm}
  \centering
  \subfigure[\emph{Recall} v.s. Time]{
  \includegraphics[width=0.22\textwidth, height=0.15\textwidth]{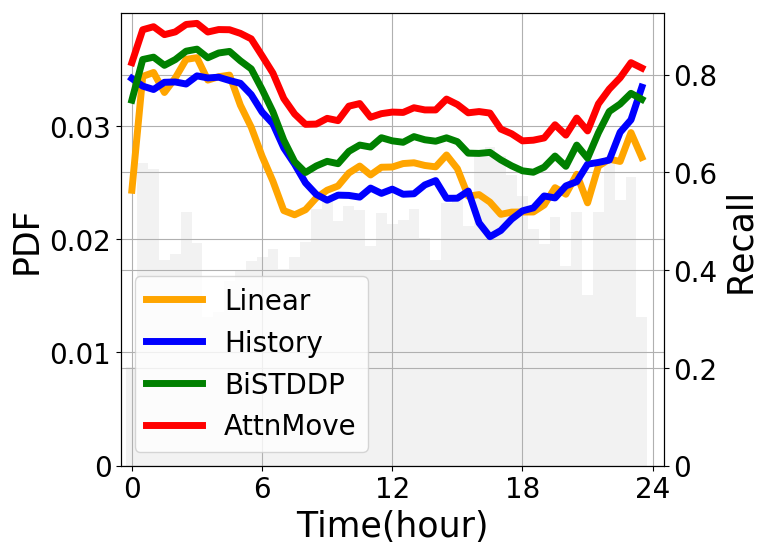}}
  \subfigure[\emph{Recall} v.s. \#location]{
  \includegraphics[width=0.22\textwidth, height=0.15\textwidth]{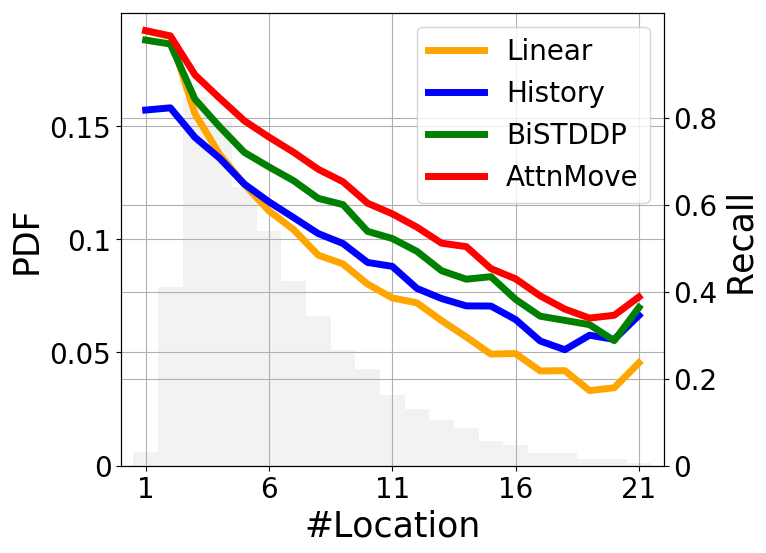}}
  \vspace{-4mm}
  \caption{Model performance in different contexts, where the solid lines denote \emph{Recall} and the shades show the distribution.}
  \vspace{-6mm}
\end{figure}

We finally investigate the sensitivity of dimension $d$, the head number $H$, and the number of layers $N$, which determine the representation ability of the model.
 Figure 5(a) presents the performance on different embedding dimension values. We can observe that as the dimension increases, the performance is gradually improved, and when it larger than 128, the performance becomes stable. This is why we select embedding size as 128. 
Then, we conduct a grid research for $H$ and $N$. Figure 5(b) partly shows the results. We find that a larger number of layers general achieves better performance while the impact of head is not significant. Considering that more layers makes the model more expressive but requires more computational cost, to make a compromise between performance and efficiency, we finally fix the number of layer and head as 4 and 8, respectively.

\begin{figure}[t]
  \centering
  \subfigure[Embedding size]{
  \includegraphics[width=0.2\textwidth]{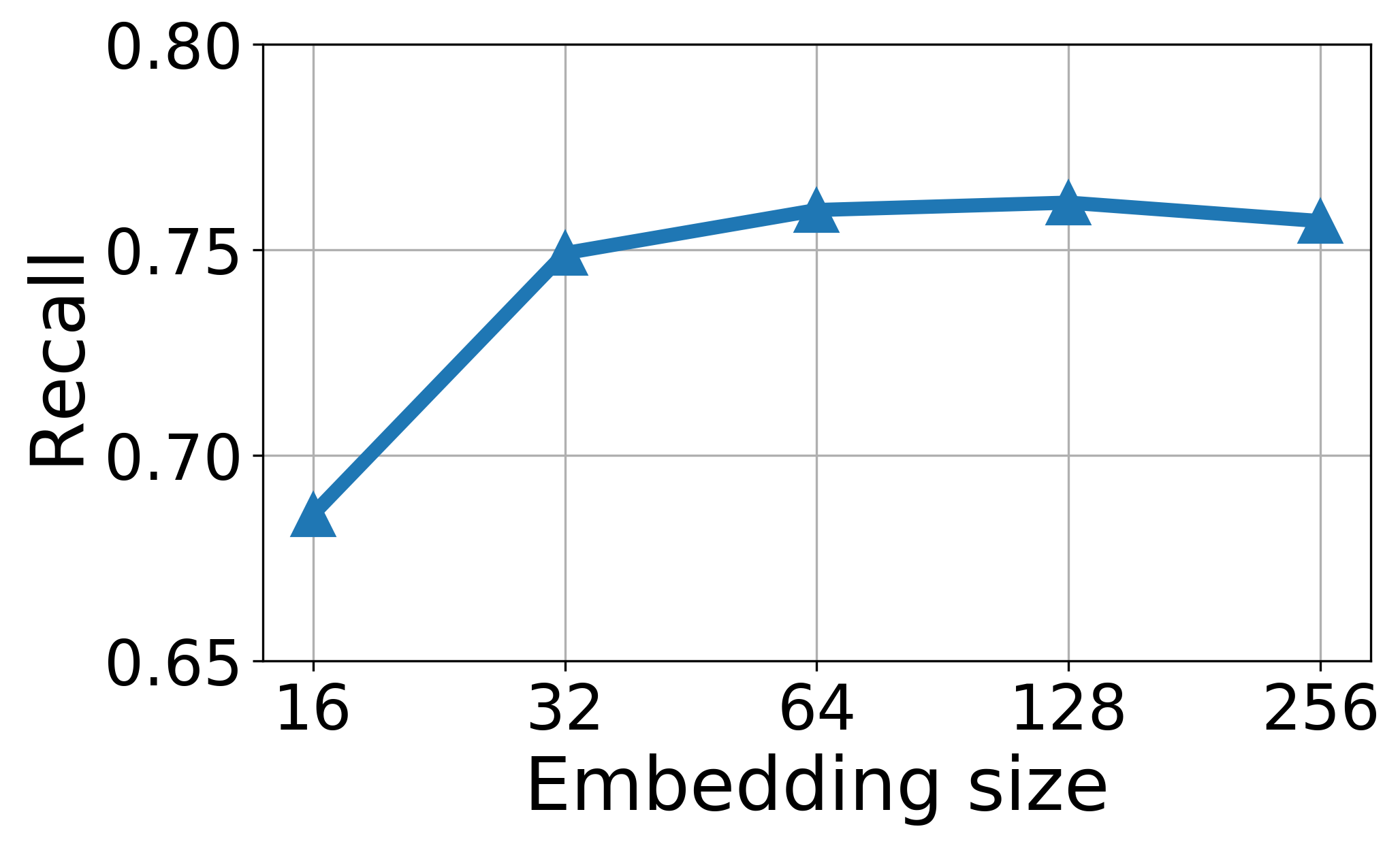}}
  \subfigure[\#Head and layer]{
  \includegraphics[width=0.2\textwidth]{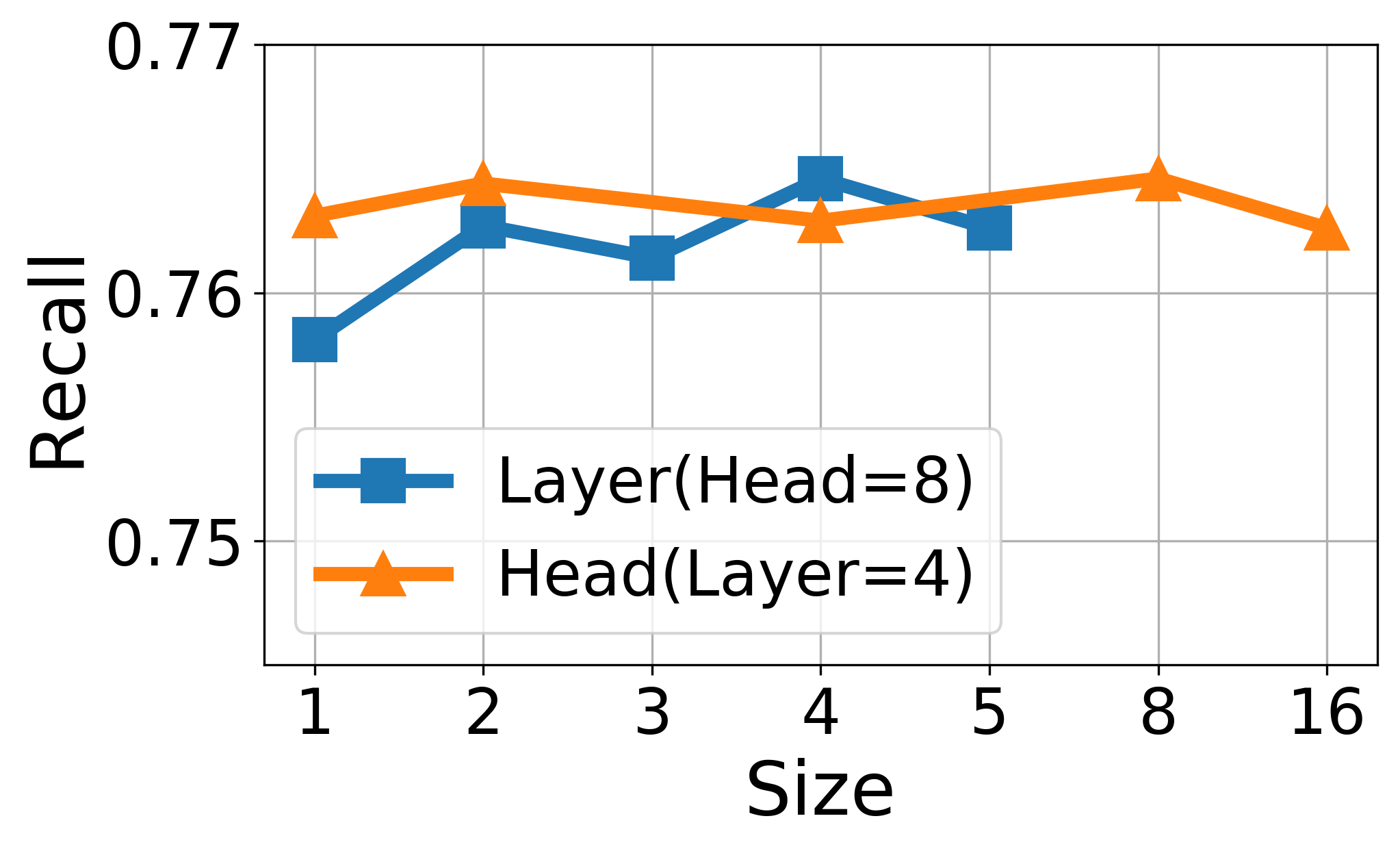}}
 \vspace{-2mm}
  \caption{Key hyper-parameter tuning.}
   \vspace{-5mm}
\end{figure}

\section{Related Work} \label{sec:relatedwork}

\textbf{Trajectory Recovery.} 
Recovering the missing value for time series is an important problem, for which deep learning models have achieved promising performance~\cite{luo2018multivariate}. However, they are insufficiently precise to be applied with mobility data due to their inability to model spatial-temporal dependence and users' historical behaviours.
Apart from these recovery methods, some works for mobility prediction can also be adopted for recovery.
Feng \emph{et al.}~\cite{feng2018deepmove} incorporated the periodicity of mobility learned from history with the location transition regularity modeled by recurrent neural network (RNN) to predict the next location. However, the performance declines because it cannot model the spatial-temporal dependence from the sparse trajectory.
As such, models for mobility data recovery have been particularly studied.
Li \emph{et al.}~\cite{feng2018deepmove} used entropy and radius such trajectory-specific features for spatial-temporal dependence modeling, while it failed to exploit history. 
 By presenting a user's long-term history as a tensor with day-of-the-month, time-of-the-day, and location-of-the-time three dimensions, a tensor factorization-based method was proposed~\cite{chen2019complete}. However, for this to work, it is required that the tensor is low-rank, and thus cannot model the randomness and complexity of mobility.
 Recently, Bi-STDDP was designed to represent users' history by a vector and combine it the spatial-temporal correlations for recovery~\cite{xi2019modelling}. However, the expressiveness of the vector is limited as it is unable to reflect the dynamic importance of history.
Overall, both the general time series recovery methods, trajectory prediction methods, and the existing mobility data recovery methods are incapable of tackling the instinctive challenges of the mobility recovery problem. By contrast, we propose an attentional neural network-based model, which can better model spatial-temporal dependence for sparse trajectory and exploit history more efficiently.
\vspace{3mm}
\\\textbf{Attentional Neural Network.} The recent development of attention models has established new state-of-the-art benchmarks in a wide range of applications.
Attention is first proposed in the context of neural machine translation~\cite{bahdanau2014neural} and has been proved effective in a variety of tasks such as question answering~\cite{sukhbaatar2015end}, text summarization~\cite{rush2015neural}, and recommender systems~\cite{he2018nais,song2019autoint}. Vaswani et al. \cite{vaswani2017attention} further proposed multi-head self-attention, renewed as \emph{Transformer}, to model complicated dependencies between words for machine translation. It makes significant progress in sequence modeling, as it uses fully attention-based architecture, which discards RNN but outperforms RNN-based models.  
Researchers also showed consistent performance gains by incorporating attention with RNN for mobility modeling, such as location prediction~\cite{feng2018deepmove}, and personalized route recommendation~\cite{wang2019empowering}, where attention can make up RNN's limitation in capturing long-term temporal dependence. Distinct from previous researchers, we are the first to adopt the fully attentional neural network, \emph{Transformer} alike, to tackle the mobility data recovery problem.

\section{Conclusion and Future Work}
In this paper, we proposed an attentional neural network-based model AttnMove to recover user's missing locations at a fine-grained spatial-temporal.
To handle the sparsity, we designed an intra-trajectory mechanism to better model the mobility regularity.
To make full use of history and distill helpful periodic characters, we proposed to integrate relatively long-term historical records.
Finally, we also designed an inter-trajectory mechanism to effectively fuse the mobility regularity and the historical pattern.
Extensive results on real-worlds datasets demonstrate the superiority of AttnMove compared with the state-of-the-arts.

In the future, we plan to extend our framework by incorporating some features like the function of location or the points of interest users visited to achieve semantic-aware trajectory interpolation.

\section{Acknowledgments}
This work was supported in part by The National Key Research and Development Program of China under grant 2020AAA0106000, the National Nature Science Foundation of China under U1936217,  61971267, 61972223, 61941117, 61861136003, Beijing Natural Science Foundation under L182038, Beijing National Research Center for Information Science and Technology under 20031887521, and research fund of Tsinghua University - Tencent Joint Laboratory for Internet Innovation Technology.  

\bibliography{main}

\end{document}